\documentclass[letterpaper]{article} 
\usepackage{aaai25}  
\nocopyright
\usepackage{times}  
\usepackage{helvet}  
\usepackage{courier}  
\usepackage[hyphens]{url}  
\usepackage{graphicx} 
\usepackage{natbib}  
\usepackage{caption} 
\usepackage{algorithm}
\usepackage{algorithmic}
\usepackage{amsmath}
\usepackage{amsfonts}

\usepackage{newfloat}
\usepackage{listings}
\DeclareCaptionStyle{ruled}{labelfont=normalfont,labelsep=colon,strut=off} 
\floatstyle{ruled}
\newfloat{listing}{tb}{lst}{}
\floatname{listing}{Listing}
\title{ConceptSearch: Towards Efficient Program Search Using \\ LLMs for Abstraction and Reasoning Corpus (ARC)}
\author{
    Kartik Singhal, 
    Gautam Shroff
}
\affiliations{
    IIIT Delhi, India
    \\
    \{kartik21259, gautam.shroff\}@iiitd.ac.in
    \thanks{This work was conducted while Kartik Singhal was an intern and Gautam Shroff was employed at TCS Research}

}

\usepackage{fancyhdr} 
\usepackage{lipsum}   

\fancypagestyle{firstpage}{
    \fancyhf{} 
    \fancyhead[L]{Pre-print of paper accepted at AAAI} 
}

\begin{document}
\thispagestyle{firstpage}

\maketitle

\begin{abstract}

The Abstraction and Reasoning Corpus (ARC) poses a significant challenge to artificial intelligence, demanding broad generalization and few-shot learning capabilities that remain elusive for current deep learning methods, including large language models (LLMs). While LLMs excel in program synthesis, their direct application to ARC yields limited success. To address this, we introduce ConceptSearch, a novel function-search algorithm that leverages LLMs for program generation and employs a concept-based scoring method to guide the search efficiently. Unlike simplistic pixel-based metrics like Hamming distance, ConceptSearch evaluates programs on their ability to capture the underlying transformation concept reflected in the input-output examples. We explore three scoring functions: Hamming distance, a CNN-based scoring function, and an LLM-based natural language scoring function. Experimental results demonstrate the effectiveness of ConceptSearch, achieving a significant performance improvement over direct prompting with GPT-4. Moreover, our novel concept-based scoring exhibits up to 30\% greater efficiency compared to Hamming distance, measured in terms of the number of iterations required to reach the correct solution. These findings highlight the potential of LLM-driven program search when integrated with concept-based guidance for tackling challenging generalization problems like ARC.
\end{abstract}

%
\begin{links}
    \link{Code}{https://github.com/kksinghal/concept-search}
\end{links}

\section{Introduction}

The Abstraction and Reasoning Corpus (ARC) constitutes a significant benchmark in artificial intelligence, specifically designed to evaluate the development of general-purpose intelligence \cite{chollet2019measureintelligence}. In contrast to other benchmarks that often prioritize pattern recognition or domain-specific expertise, ARC emphasizes fundamental cognitive skills, including abstraction, reasoning, and generalization. The corpus comprises a set of analogy puzzles, each presenting a series of input-output pairs (typically 2-4) that embody a latent transformation rule or concept. The central challenge lies in inferring this underlying transformation rule and subsequently applying it to previously unseen test input.

The examples consist of an "input grid" and an "output grid," each featuring 10 symbols (visualized as unique colors) and its size ranges from 1$\times$1 to 30$\times$30 in size. To solve an evaluation task, a test-taker uses the provided training examples and input grid of the test example to construct the output grid from scratch, determining its dimensions and symbol placement. Success is binary, achieved by correctly predicting the output grid for all test examples in a task, with up to three attempts. An intelligent system's performance on ARC is the fraction of tasks it successfully solves, measuring "developer-aware generalization," with no prior knowledge of evaluation set tasks assumed.

\begin{figure}[t]
\centering
\includegraphics[width=0.9\columnwidth]{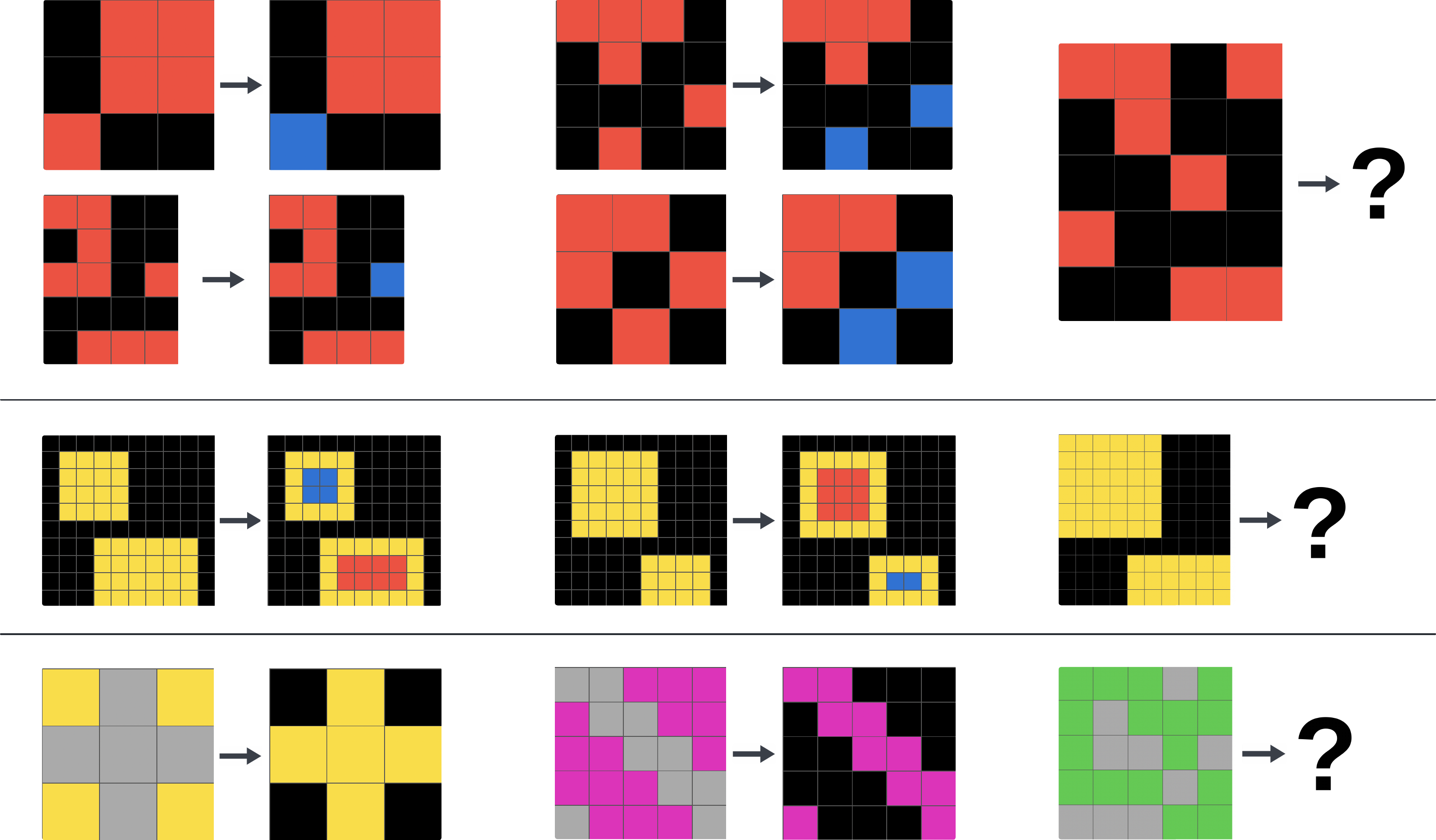}
\caption{Three sample ARC tasks, easily solvable by humans, yet unsolved by our proposed method as well as GPT-4 baseline \cite{xu2024llmsabstractionreasoningcorpus}}
\label{unsolved_tasks}
\end{figure}

Even though humans can solve most of the tasks, to the best of our knowledge, no current machine learning techniques, including Deep Learning, are well-suited to tackle the ARC benchmark (see Figure \ref{unsolved_tasks}). This is due to ARC's emphasis on broad generalization and few-shot learning, coupled with the fact that each task is unique, therefore the evaluation set consists of concepts that are not present in the training set.

Since purely neural approaches, including large language models (LLMs), often fail to produce correct output grid in end-to-end manner \cite{mitchell2023comparinghumansgpt4gpt4v, boberirizar2024neuralnetworksabstractionreasoning, xu2024llmsabstractionreasoningcorpus}, most of the current top approaches frame it as a program-synthesis problem. This strategy avoids the black-box nature of neural models and allows for high expressivity and the use of search-based methods. Instead of using a highly open-ended programming language such as Python, these methods employ hand-crafted languages, known as Domain-Specific Languages (DSLs). A DSL is designed to ensure that all ARC tasks can be solved using it, while being abstract and generic by defining a small number of versatile primitives, each applicable to numerous ARC tasks.

Most approaches can be broadly classified into three categories: brute-force search. neural-guided search and LLM-based techniques. Brute-force search and neural-guided search-based methods require careful hand-crafted DSL, but can still scale poorly to complex problems due to combinatorial complexity. LLM-based techniques aim to either generate the output grid directly or generate a program that transforms the input grids to output grids without any feedback loop.

\cite{redwoodresearch} has demonstrated that sampling a large number of programs using GPT-4o leads to impressive performance on the ARC-AGI benchmark, exhibiting a scaling law between the number of samples generated and the number of tasks solved. Although this approach is computationally demanding, it highlights the potential of LLMs to generate solution programs for these tasks.

FunSearch \cite{funsearch} proposed a function-search algorithm for problems in mathematical sciences utilizing an LLM to iteratively evolve programs within its database. At each iteration, the LLM takes two programs sampled from the database, ranked according to a predefined scoring function. Taking inspiration from these sampled programs and leveraging their relative scores as indicators of proximity to the desired solution, the LLM generates a new, potentially improved program. This iterative process, driven by LLM-based program evolution, aims to converge towards increasingly accurate solutions.

FunSearch is suited for problems with efficient evaluator for determining success and a rich scoring feedback quantifying the improvements, instead of binary signal. For ARC-AGI, evaluator is simply a pixel-wise comparison between the predicted output grid and solution output grid. The challenge is the scoring function. In our problem, success is a binary measure, whether all the pixels in the predicted output grid match the true output grid. So, we need to develop a scoring function that can provide rich and useful signals to the LLM to guide the search.

A trivial scoring function is Hamming distance between the predicted output grid and true output grid, that is, the number of pixels not matching between the two grids, normalized by the size of the grid. However, relying solely on Hamming distance to evaluate programs can be misleading, as superficial resemblance can hide major differences in the program's logic and functionality. Even though visual similarity and low Hamming distance between the grids might suggest resemblance, the underlying program functions used to generate them could differ significantly from the true solution program.

Therefore, we need our scoring function to capture the concept or logic of the transformation. Our work focuses on employing FunSearch on ARC-AGI and introduces two novel concept-based scoring functions integrated with a feedback loop. These scoring functions leverage two distinct modalities: one based on vision and the other on natural language. The results demonstrate a substantial improvement in task-solving performance using FunSearch, increasing number of successfully solved tasks from 13/50 to 25/50. Our concept-based scoring function further improves the task success to 29/50 while significantly enhancing the efficiency of the function search by $\sim$30\% compared to using the Hamming distance.

\section{Related Works}
\textbf{Brute-force search.} 
The winner of Kaggle ARC Challenge 2020 \cite{icecuber} implements a DSL with 142 hand-crafted unary functions on grids. At runtime, the functions are greedily composed on the input grids, with the resulting ‘pieces’ stored in a directed acyclic graph (DAG). Finally, a solver combines pieces from the DAG to get as close as possible to matching the training examples.

\cite{arga} implements a constrain-guided search by converting ARC grids into an object-graph representation and operating on these representations using a graph-based DSL. However, it only works for the same input-output grid size.

\cite{ainooson2023neurodiversityinspiredsolverabstraction} proposed a DSL called Visual Imagery Reasoning Language (VIMRL) based on core knowledge of objectness for multi-level search-based program synthesis, allowing flexible use of the core knowledge to address ARC tasks.

\textbf{Neural-guided search.} \cite{boberirizar2024neuralnetworksabstractionreasoning} adapts the Dreamcoder framework \cite{ellis2020dreamcodergrowinggeneralizableinterpretable} designed to grow interpretable and generalizable knowledge through wake-sleep Bayesian program learning consisting of iterative phases to improve program synthesis, without the abstraction phase of growing the underlying library of code primitives.

HYSYNTH \cite{barke2024hysynthcontextfreellmapproximation} uses an LLM to generate sample programs and learn a probabilistic context-free grammar (PCFG). This learned PCFG is then used to guide the bottom-up search for program synthesis.

\textbf{LLM-based approaches.} 
\cite{boberirizar2024neuralnetworksabstractionreasoning} compare multiple LLMs for ARC by directly predicting the output grid based on provided demonstration examples. The results reveal a substantial enhancement of $\sim$2x in performance from GPT-3.5 to GPT-4. 

\cite{mitchell2023comparinghumansgpt4gpt4v} evaluate the performance of both text-only and multimodal GPT-4 on the ConceptARC benchmark \cite{conceptarc} and conclude that neither version of GPT-4 has developed robust abstraction abilities at human-like levels.

Hypothesis search \cite{wang2024hypothesissearchinductivereasoning} aims to decouple program generation into explicit hypothesis generation in natural language (NL) and program generation from NL hypothesis. Natural language offers abstract but ambiguous representations. Programmatic hypotheses, though detailed and verifiable, may distract language models. Their results show that explicit hypothesis formation significantly outperforms direct prompting.

\begin{figure*}[t]
\centering
\includegraphics[width=0.8\textwidth]{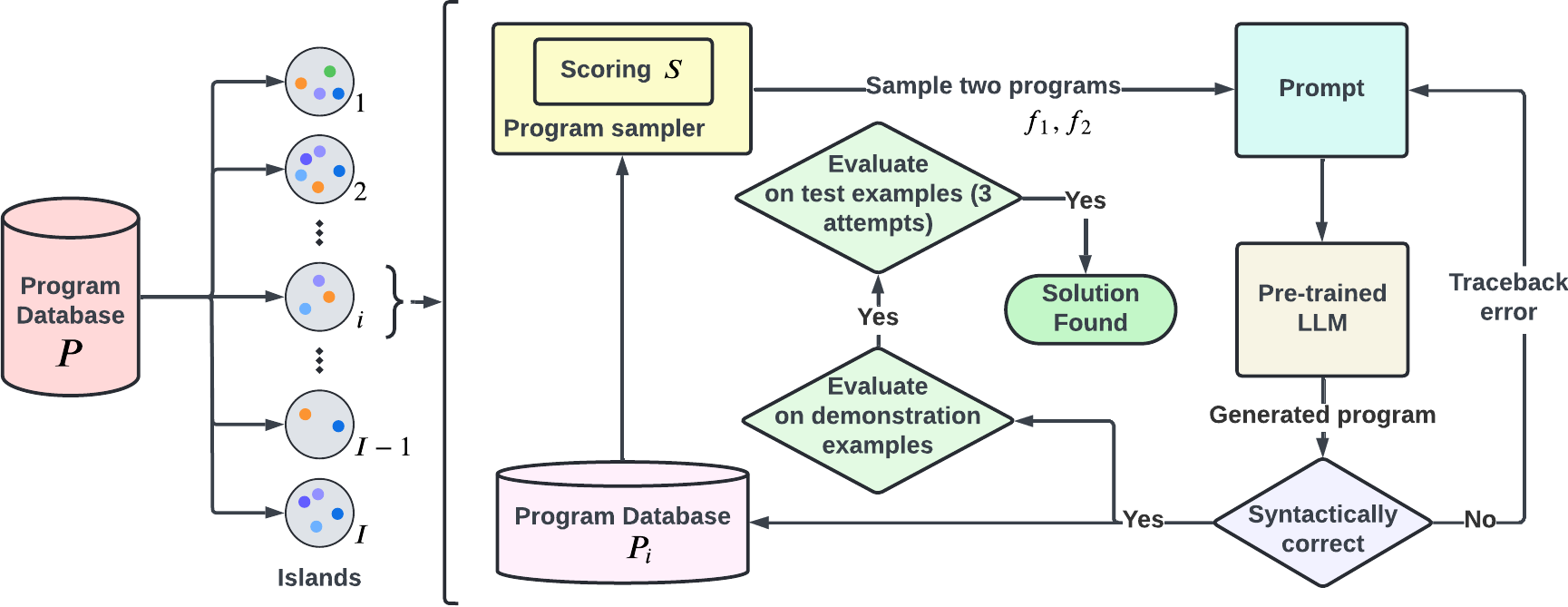} 
\caption{Flowchart of the function-search algorithm, illustrating how programs in program database $P$ are evolved using scoring function $S$ in context of Abstraction and Reasoning Corpus.}
\label{fig2}
\end{figure*}

\cite{xu2024llmsabstractionreasoningcorpus} shows that the main issue is the LLM's difficulty in maintaining "object cohesion" across ARC image grids due to the grids' two-dimensional nature. To address this, they introduced a 1D-ARC benchmark, a set of ARC-like tasks represented as a single line of text. It was found that GPT performs better on 1D-ARC tasks than on standard ARC tasks. Using object-centric graph abstractions from \cite{arga} for input-output grids significantly improves GPT-4's performance.

Code-Iteration \cite{codeit} aims to learn a code-generation policy (T5 model) through hindsight-relabelling, which involves learning from its own generated programs for inputs and realised outputs for those programs. This is coupled with growing DSL primitives through a mutation algorithm.

\section{Methodology}

A task $\tau = [(I_{1:n}, O_{1:n}), (I_{{t,}_{1:n'}}, O_{{t,}_{1:n'}})]$ consists of set of $n$ demonstration examples $(I_{1:n}, O_{1:n})$ exhibiting a common latent transformation. The goal is to infer that transformation and apply it to each of $n'$ test input grids $I_{{t,}_{1:n'}}$ to get the corresponding test output grids $O_{{t,}_{1:n'}}$.

\subsection{ConceptSearch}

The algorithm begins with a program database $P$ containing initial programs. The objective is to generate candidate solutions using a pre-trained LLM leveraging programs from $P$ for in-context learning. These two programs are selected using a scoring function $S$, based on a similarity measure. Each newly generated program is evaluated to determine whether it solves the task $\tau$. Otherwise, it is added back to the program database. By iterating through this process, the program database "evolves" into a repository of new knowledge, and the LLM eventually generates the solution program. 

In order to make the search faster, $I(=5)$ independent experiments, called islands, are run in parallel. Each of the islands has its own program database $P_i$ initialized with the top-2 programs in the program database $P$ using the scoring function $S$.

\begin{figure}[t]
\centering
\includegraphics[width=0.99\columnwidth]{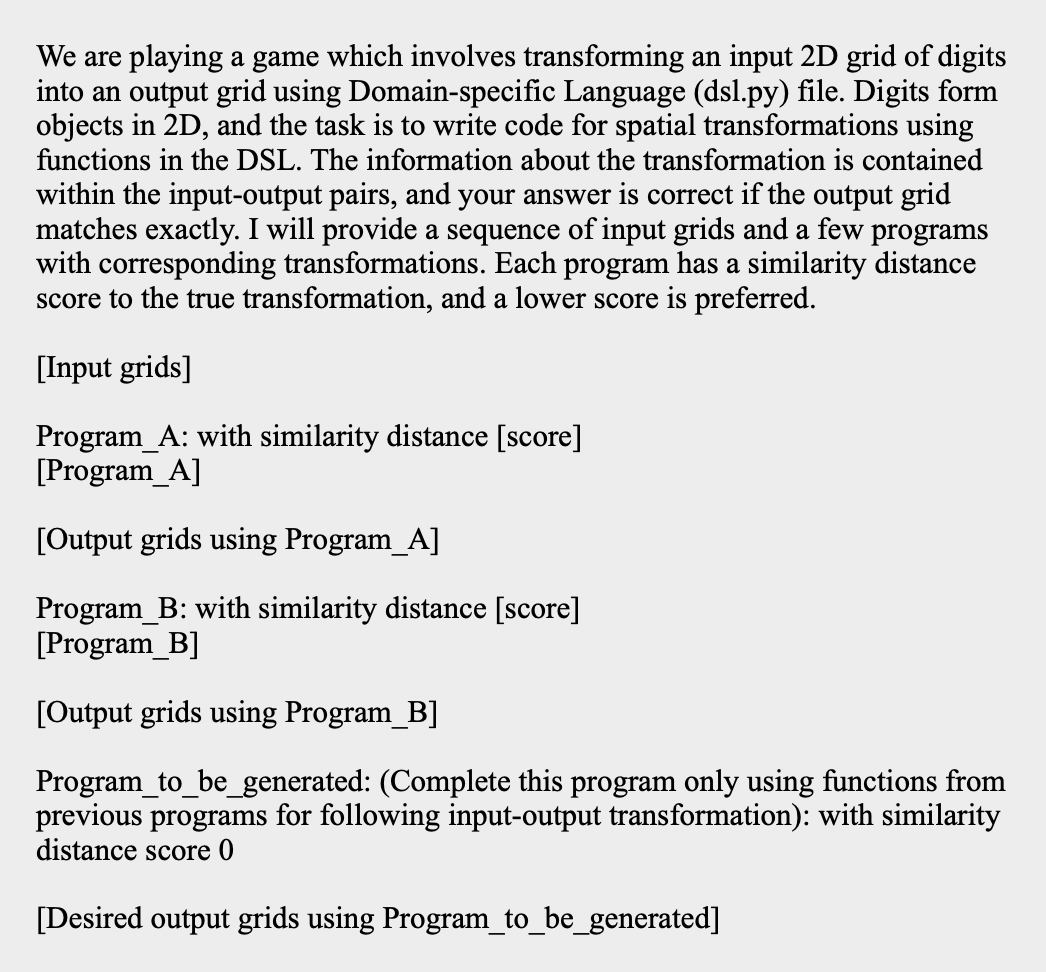}
\caption{Compact version of prompt used in program-generation step with two in-context program examples}
\label{program_generation_prompt}
\end{figure}

We adopt the ARC-DSL \cite{arc-dsl} as our Domain-Specific Language and the provided program solvers of training tasks to initialise the program database. At each of the program-generation step in island $i$, two programs $f_1$ and $f_2$ are sampled from $P_i$ using a probability distribution based on the scoring function $S$ (more details on $S$ in later sections). The prompt consists of initial problem context, input grids $I_{1:n}$ of demonstration examples, similar programs $f_1$ and $f_2$ along with similarity scores, their realised outputs $f_1(I_{1:n})$ and $f_2(I_{1:n})$, and then finally the desired outputs $O_{1:n}$ of the task (see Figure \ref{program_generation_prompt}). Similarity scores indicate which program is nearer to the solution, thereby offering more guidance to the LLM. Additionally, the program definitions of ARC-DSL in Python are provided for a better understanding of each function.

\renewcommand{\algorithmicrequire}{\textbf{Input Task:}}
\begin{algorithm}[t]
\caption{ConceptSearch algorithm}
\begin{algorithmic}[1]
\REQUIRE $(I_{1:n}, O_{1:n}), (I_{{t,}_{1:n'}}, O_{{t,}_{1:n'}})$ \\

\STATE $S$ = \textit{ScoringFunction()} 
\STATE program\_db = \textit{ProgramDatabase()}

\FOR{iteration in range(island\_iterations)}
    \STATE $f_{i1}$, $f_{i2}$ = \textit{program\_db.get\_2\_closest\_fs($I_{1:n}$, $O_{1:n}$, $S$)}
    
    \STATE island = \textit{Island($f_{i1}$, $f_{i2}$)}

        \FOR{step in range(program\_generation\_steps)}
            \STATE $f_1$, $f_2$ = \textit{island.sample\_2\_closest\_fs($I_{1:n}$, $O_{1:n}$, $S$)}
            
            \STATE f = \textit{gen\_program($f_1$, $f_2$, $I_{1:n}$, $O_{1:n}$, $S$)}
            
            \STATE (syntactically\_correct, error) = \textit{run(f)}

            \IF{\textit{syntactically\_correct(f)}}
                \STATE \textit{island.add\_program(f)}
                \IF{\textit{evaluate(f, $I_{1:n}$, $O_{1:n}$)} == 1}
                    \IF{\textit{evaluate(f, $I_{{t,}_{1:n'}}$, $O_{{t,}_{1:n'}}$)} == 1}
                        \RETURN \textit{"found\_solution"}
                    \ELSIF{num\_attempts $\geq$ 3}{
                        \RETURN \textit{"solution\_not\_found"}
                    }
                    \ENDIF
                \ENDIF
            \ELSE
                \STATE \textit{island.add\_syntax\_error\_to\_next\_prompt(error)}
            \ENDIF
    \ENDFOR

    \STATE \textit{program\_db.add\_newly\_found\_programs(island)}
\ENDFOR

\RETURN \textit{"solution\_not\_found"}

\end{algorithmic}
\end{algorithm}

In each of these program-generation steps, 5 different programs are generated in a single response. All of these generated programs are run with training grids $I_{1:n}$ as input and are evaluated by comparing the dimensions and pixel equivalence with $O_{1:n}$. In case of a syntax error, a feedback loop adds the traceback error in the prompt of the next iteration to potentially fix that syntax and generate the intended syntactically correct program. If a program is found that generates correct output grids for all demonstration examples, that program is submitted for evaluation with test input-output grids $(I_{{t,}_{1:n'}}, O_{{t,}_{1:n'}})$. If it is a success, then the task is solved. Otherwise, the algorithm continues for a maximum of 3 evaluation attempts for the test grids.

These function-generation steps are iterated 10 times, after which all the new programs in $P_{1:I}$ are added to $P$ and all islands are reinitialized to allow sharing of knowledge gained across multiple islands. This process of island-iteration is performed twice. These numbers are chosen for reasonable compute cost and can be scaled up for better performance \cite{redwoodresearch}. In our work, the maximum number of API calls to the LLM for a single task is given by $I \times$ program-generation-steps $\times$ island-iterations = $5 \times 10 \times 2 = 100$.

We hypothesize that the quality of the programs provided in context directly influences the efficiency of the search process and show that the choice of scoring function $S$ is fundamental to solving complex tasks in a reasonable compute and time, rather than sampling a large number of programs. 

Our objective is to create a function that maps input-output grids into a rich embedding space, effectively encoding the transformation conceptually. This will assist LLMs in selecting the appropriate DSL functions for the transformation demonstrated in the examples. A scoring function will then calculate the Euclidean distance between the latent vectors of the transformations in the demonstration examples and that of the given program. Therefore, the score will always be non-negative, and the program with the smallest score will be the one most closely aligned with the transformation demonstrated in the examples.

For sampling programs using the scores, the probability of choosing a program $f$ is calculated as:
\begin{equation} \label{scores_to_probability}
p(f) = \frac{e^{-S(f)}} {\displaystyle\sum_{f_i \in P} e^{-S(f_i)}}
\end{equation}

\subsection{CNN-based Scoring Function}\label{sec:cnn_scoring_function}

For demonstration examples $(I_{1:n}, O_{1:n})$ and a given program \(f\), we first compute the realized outputs $f(I_{1:n})$. This yields two sets of transformations: the input-output grid pairs from the demonstration examples $(I_{1:n}, O_{1:n})$ and the input-output pairs produced by the program $(I_{1:n}, f(I_{1:n}))$. Our goal is to compute the distance between these two sets of transformations.

\begin{figure}
\centering
\includegraphics[width=\columnwidth]{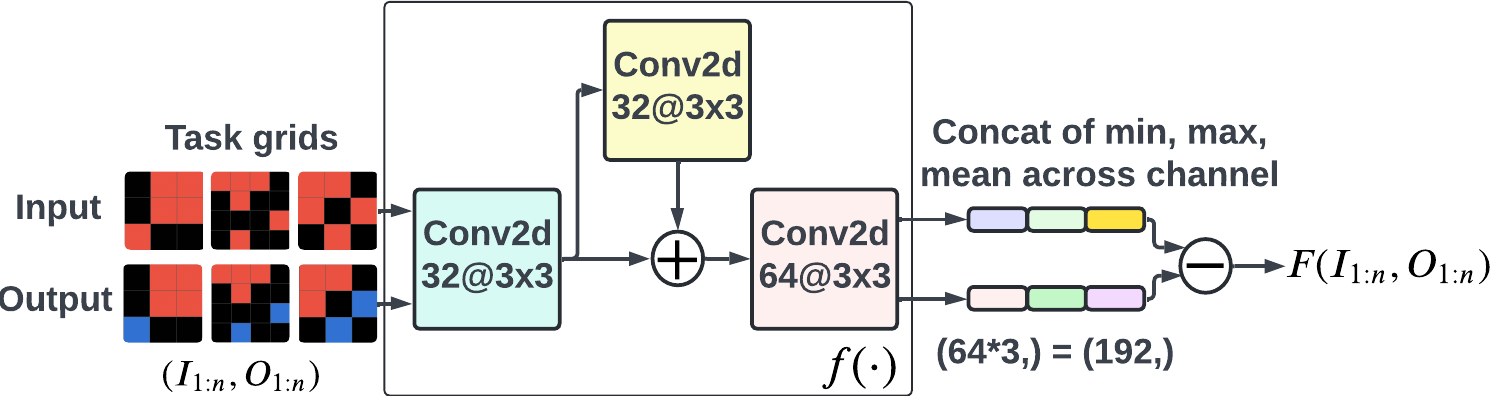}
\caption{Model architecture trained with classification loss and contrastive loss for learning meaningful task representations. The grid size can be as small as 1$\times$1, and cell occupancy is one-hot encoded into 10 channels, denoting 9 different colours and one for the cell being empty.}
\label{cnn_model}
\end{figure}

To train a neural network capable of capturing the concept of transformation, we utilize the Concept-ARC dataset \cite{conceptarc}. This dataset classifies various tasks into 16 distinct classes, with each class containing 11 manually created ARC tasks. In total, the dataset comprises 176 tasks, systematically categorized based on specific concepts. Our neural-network architecture is inspired by \cite{boberirizar2024neuralnetworksabstractionreasoning} for handling variable-sized grids (see Figure \ref{cnn_model}). The minimum grid size is 1$\times$1, therefore, the convolution layers keeps the grid size same. Dilated convolution layer \cite{yu2016multiscalecontextaggregationdilated} is also used, which create gaps between sampled pixels, mimicking a convolution on a downsampled image. 

Given the variability in grid sizes, it is necessary to use an aggregator function for each channel to generate a feature vector of constant size that is independent of the grid size. This ensures that grids with different sizes can be represented in the same feature space. Specifically, we compute the minimum, maximum, and mean of each channel, then concatenate these values to form a flattened feature vector.

For each input-output pair in the task grids, the pair is passed through the model. The mean of the feature vectors for each pair, after performing a difference operation, is calculated. The final feature vector $z$, which integrates information from all input-output pairs, effectively represents the underlying transformation rule (see Algorithm \ref{alg:cnn_model_algo}). This resultant vector is then processed through the classification and projection layers.

This model $F$ is trained using a dual loss approach, incorporating both cross-entropy loss and contrastive loss (triplet-margin loss) \cite{khosla2021supervisedcontrastivelearning} to classify tasks into 16 distinct concept classes. The contrastive loss is applied on the projection layer and ensures that samples within the same class are closer together in the feature space, while samples from different classes are pushed farther apart. The feature vector $z=F((I, O))$, the mean of the difference vectors, serves as the embedding vector, allowing a meaningful representation of the transformation concept.

The number of model parameters is kept small due to the small size of the Concept-ARC dataset (176 samples) to avoid overfitting. For better generalization, data augmentation techniques were used such as rotation, transpose of both input-output grids and random permutation of colors across all the task examples. Since the dataset is small, we use k-fold cross-validation (k=5) for hyperparameter tuning and take the mean of feature vectors obtained from each of the 5 models.

The similarity score between the transformation underlying the demonstration examples $(I_{1:n}, O_{1:n})$ and a given program $f$ is computed as:
\begin{equation} \label{eq1}
\begin{split}
S_{\text{CNN}}((I_{1:n}, O_{1:n}), f) = S_{\text{CNN}}((I_{1:n}, O_{1:n}), (I_{1:n}, f(I_{1:n}))) & \\
 = \lVert F(I_{1:n}, O_{1:n}) - F(I_{1:n}, f(I_{1:n})) \rVert &
\end{split}
\end{equation}

\renewcommand{\algorithmicrequire}{\textbf{Input:}}
\renewcommand{\algorithmicensure}{\textbf{Output:}}
\begin{algorithm}[t]
\caption{Inference of transformation embedding vector from task grids using CNN-based model (figure. \ref{cnn_model})}
\label{alg:cnn_model_algo}
\begin{algorithmic}[1]
\REQUIRE
$\mathcal{D} = (I_{1:n}, O_{1:n})$, a set of input-output grid pairs representing a latent transformation
\ENSURE A transformation embedding vector $z \in \mathbb{R}^{192}$ representing the transformation embedding.
\STATE $\mathcal{D}_d \leftarrow \emptyset$ \COMMENT{Initialize an empty list for difference vectors}
\FOR{$(I_i \in \mathbb{R}^{10 \times h_i \times w_i}, O_i \in \mathbb{R}^{10 \times h_o \times w_o})$ in $\mathcal{D}$}
    \STATE \textbf{Feature Extraction:}
    \STATE $F_I \leftarrow f(I_i) \in \mathbb{R}^{64 \times h_i \times w_i}$
    \STATE $F_O \leftarrow f(O_i) \in \mathbb{R}^{64 \times h_o \times w_o}$
    \STATE \textbf{Channel-wise Aggregation:} along $h \times w$
    \STATE $v_I \leftarrow [\text{min}(F_I), \text{max}(F_I), \text{mean}(F_I)] \in \mathbb{R}^{192}$
    \STATE $v_O \leftarrow [\text{min}(F_O), \text{max}(F_O), \text{mean}(F_O)] \in \mathbb{R}^{192}$
    \STATE \textbf{Difference Vector Calculation:}
    \STATE $d_i \leftarrow v_I - v_O$
    \STATE $\mathcal{D}_d \leftarrow \mathcal{D}_d \cup \{d_i\}$
\ENDFOR
\STATE \textbf{Embedding Generation:}
\STATE $z \leftarrow \frac{1}{n} \sum_{d_i \in \mathcal{D}_d} d_i$
\RETURN $z$
\end{algorithmic}
\end{algorithm}

\begin{figure}[t]
\centering
\includegraphics[width=0.9\columnwidth]{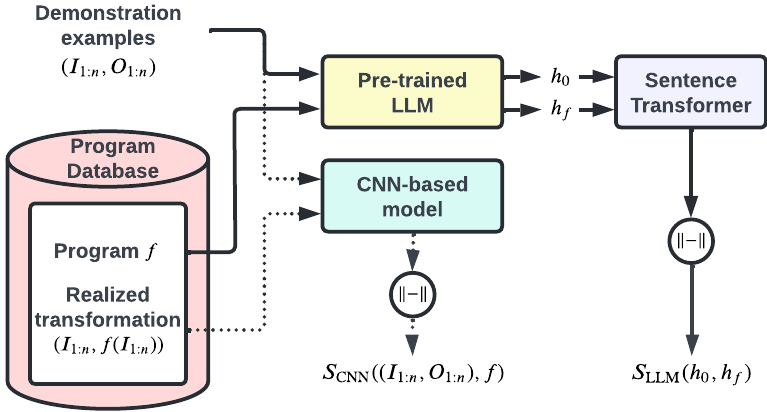}
\caption{Comparing CNN-based and LLM-based scoring: one extracts features via CNN, while the other leverages LLM for natural language hypothesis, which is then converted to feature embedding by using SentenceTransformer.}
\label{scoring_function_overview}
\end{figure}

\subsection{LLM-based Natural Language Scoring Function}

In the previous scoring function, the feature extractor was a convolutional neural network. In this scoring function, we would like to use a pre-trained LLM for feature extraction. Therefore, for each transformation $(I_{1:n}, O_{1:n})$, we generate a transformation hypothesis $h_0$ in natural language (NL) using an LLM. Then, a text embedding model maps this natural language hypothesis into rich embedding, which can be again used for calculating the Euclidean distance between two transformations.

\begin{figure}[t]
\centering
\includegraphics[width=0.98\columnwidth]{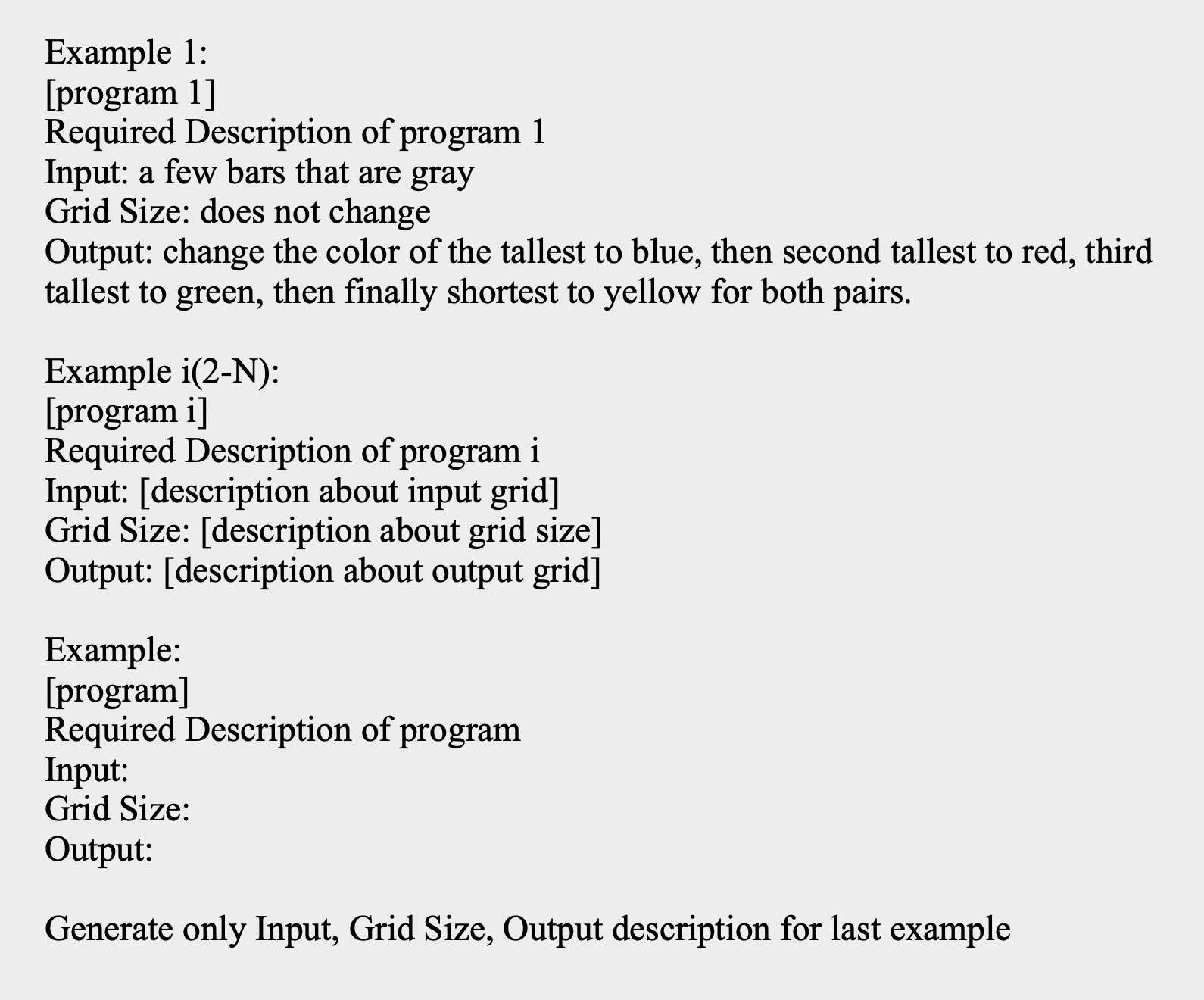}
\caption{Compact version of prompt used in hypothesis-generation step for generating natural language description of transformation underlying the given program with N in-context examples}
\label{hypothesis_generation_prompt_for_program}
\end{figure}

LARC \cite{larc} is a dataset consisting of natural language descriptions on how to solve each task provided by human participants. These descriptions are only available for 354 tasks in the training set, where a human participant was successfully able to solve the task using a natural language description of the transformation provided by another human participant. For the rest of the programs and each new LLM-generated program added to the program database, a description is generated using a pre-trained LLM as a completion task by providing existing programs in the program database and their descriptions for in-context learning (see Figure \ref{hypothesis_generation_prompt_for_program}). Additionally, the program definitions of ARC-DSL in Python are provided for a better understanding of each function.

For solving a task with demonstration examples $(I_{1:n}, O_{1:n})$, our objective is to find a natural language hypothesis that corresponds to these examples. This hypothesis will serve as a reference point from which we calculate the distance to other candidate hypotheses. To generate this "goal" hypothesis $h_0$, we employ a completion task.

\begin{figure}[t]
\centering
\includegraphics[width=0.98\columnwidth]{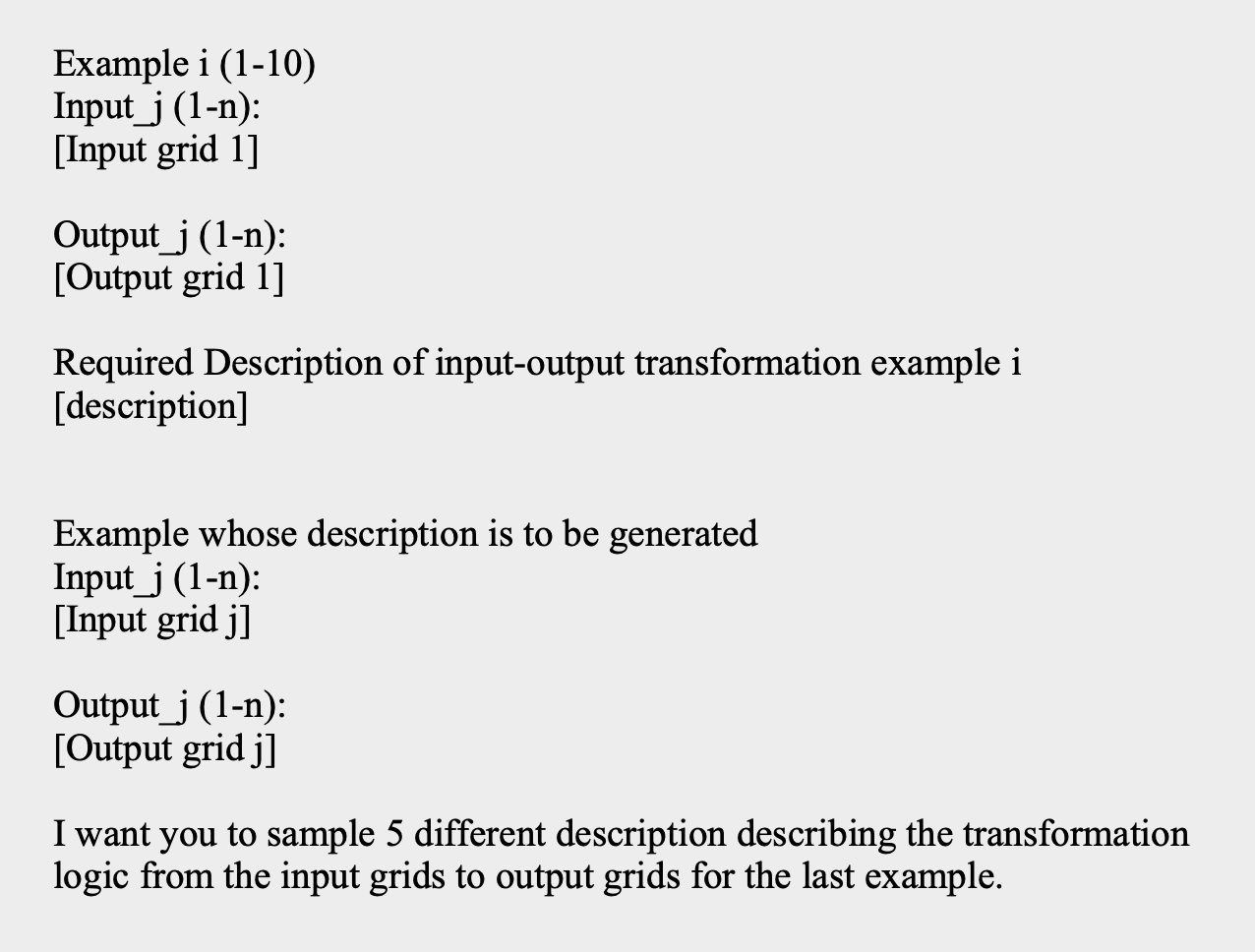}
\caption{Compact version of prompt used in goal hypothesis generation step for generating natural language description of transformation underlying input-output grids with ten in-context examples}
\label{hypothesis_generation_prompt_for_IO}
\end{figure}

We select the top-10 programs $f_i$ from our program database using a CNN-based scoring function (section \ref{sec:cnn_scoring_function}). For these selected programs, we provide their demonstration examples $(I_{1:n}, f_i(I_{1:n}))$ along with their corresponding descriptions for in-context learning (ICL). The LLM then completes the description for the task demonstration examples $(I_{1:n}, O_{1:n})$ (see Figure \ref{hypothesis_generation_prompt_for_IO}). To manage the context length effectively, we limit the in-context learning to 10 examples. 

Since we know that deep learning methods don't work well for ARC tasks including LLMs \cite{wang2024hypothesissearchinductivereasoning}, the natural language descriptions generated by LLMs may not be accurate. For this reason, we generate a unique "goal" hypothesis for each of the islands to increase the odds of finding the solution within given iterations, which is not possible with a CNN-based scoring function.

At this stage, we have natural language hypotheses, $h_0$ for the desired transformation (goal hypothesis) and $h_f$ for each of the programs in the program database. To derive a feature vector for a natural language hypothesis, we fine-tune a SentenceTransformer $F$ (all-mpnet-base-v2) \cite{reimers-2019-sentence-bert} using the ConceptARC dataset with contrastive loss (BatchAllTripletLoss). This fine-tuning process with contrastive loss ensures that conceptually similar hypotheses are positioned closer to each other in the embedding space, while conceptually different hypotheses are pushed further apart.

The similarity score between the transformation underlying the goal hypothesis $h_0$ and program description $h_f$, 
\begin{equation} \label{eq2}
S_{\text{LLM}}(h_0, h_f) = \lVert F(h_0) - F(h_f)) \rVert
\end{equation}

\section{Experiments and Results}
The evaluation set consists of 50 tasks from the training set, same as \cite{xu2024llmsabstractionreasoningcorpus}. This evaluation set of 50 tasks is chosen to limit the cost of LLM inference, and this also allows us to compare our results presented in \cite{xu2024llmsabstractionreasoningcorpus}. It is also made sure that these 50 tasks are not available in any way, including the program database, in-context training examples and training the scoring function.

In our experiments, we utilize Gemini 1.5 Pro for both program generation and goal hypothesis generation. This choice of using Gemini is made after manually experimenting with multiple ARC tasks using different LLMs. For the hypothesis generation of program definitions, we employ Gemini 1.5 Flash, which allows access to a higher number of tokens per minute. 

FunSearch \cite{funsearch} does not use much context about the problem, it should be viewed as a generator of diverse, syntactically correct programs with occasionally interesting ideas, which eventually solves the problem at hand. In our case, we provide the exact problem context; therefore, the generation is strongly conditioned. With temperature 0, it does not lead to much changes to the code upon iteration. Therefore, for program generation, the temperature is set to 1 for generating creative solutions.

In our work, we utilize 5 islands with 10 program iterations each. The islands are reset twice, resulting in a maximum number of program generations per task calculated as $I \times \text{program-generation-steps} \times \text{island-iterations} = 5 \times 10 \times 2 = 100$. During each API call for program generation, we prompt the LLM to generate 5 new programs with different logic. This strategy allows us to generate a larger number of programs while keeping the inference cost relatively low. Consequently, the maximum number of programs generated for a single task is $100 \times 5 = 500$. This approach ensures a diverse set of candidate programs for each task, enhancing the likelihood of finding optimal solutions. In instances where the LLM's response does not match the expected output, such as failing to generate any program code, the same prompt is repeated until at least one program code is produced. 

We evaluate our method using three different scoring functions: Hamming distance, CNN-based scoring, and LLM-based natural language scoring. The results are compared in Table \ref{tab:methods_accuracy} and \ref{tab:methods_efficiency}. The Hamming distance between two grids is calculated as the number of mismatched pixels normalized by the grid size. Our findings reveal a substantial improvement in performance when transitioning from direct-grid-based prompting to a function-search algorithm with Hamming distance. Specifically, task success increased from 13/50 to 25/50. This improvement shows that the function-search algorithm is effective in guiding the LLM towards the solution.

\begin{table}[t]
\centering
\begin{tabular}{|l|c|c|}
\hline
\textbf{Method} & \textbf{Accuracy} & \textbf{Mean iters} \\
\hline
\cite{xu2024llmsabstractionreasoningcorpus} - GPT-4 & 13/50 & - \\ 
\hline
Ours - Hamming distance & 25/50 & 3.70\\
Ours - CNN-based & 25/50 & 2.80\\
\textbf{Ours - LLM-based} & \textbf{29/50} & \textbf{2.05} \\
\hline
\end{tabular}
\caption{Comparison of our approach to direct-grid based prompting with GPT-4. Accuracy is the percentage of tasks solved in the evaluation set. Mean iterations is the average number of program-iterations taken to find the solution per task, considering only the tasks solved by all the methods}
\label{tab:methods_accuracy}
\end{table}

\begin{table}[t]
\centering
\begin{tabular}{|l|c|c|}
\hline
& \textbf{LLM-based} & \textbf{CNN-based} \\
\hline
\textbf{Hamming distance} & 24.7 & 29.3 \\
\hline
\textbf{CNN-based} & 10.3 & - \\
\hline
\end{tabular}
\caption{Efficiency (\%) improvement of the scoring function in each column compared to the scoring function in the corresponding row based on the number of program-iterations taken to find the solution for tasks solved by both methods.}
\label{tab:methods_efficiency}
\end{table}

However, this scoring function may be misleading in some cases. For instance, in transformations where changes from the input to the output grid are minimal, the Hamming distance is low but program may fail to capture the complexity of the actual transformation and may be far from the solution program.  Knowing that function-search is effective in solving these tasks, the objective with concept-based scoring functions is to make search more efficient with more effective guidance. The efficiency is compared based on the number of iterations it took to arrive at the solution.

For the two concept-based scoring functions, the main difference lies in the feature extraction process from the demonstration examples. The CNN-based scoring function utilizes a convolutional neural network for feature extraction. In contrast, the LLM-based method extracts features in the form of natural language hypotheses using an LLM. These natural language hypotheses are then mapped to a rich feature space using a text-embedding model. 

The performance remained consistent when using a CNN-based scoring function, achieving 25/50. However, the search was 29.3\% more efficient using CNN-based scoring function over Hamming distance, considering the intersection of tasks solved by both of them. An improvement in task success was observed with the LLM-based NL scoring function, which achieved a score of 29/50 with 10.3\% more efficient search over CNN-based scoring function. This suggests that LLMs have better feature extraction capabilities and overall effectiveness in handling ARC tasks.

A natural question arises regarding whether the performance can be further improved by combining CNN and LLM-based scoring functions. However, it was found that there is no task that the CNN-based scoring function solves that the LLM-based scoring function does not. Therefore, combining both scoring functions may not enable additional tasks to be solved; whether it improves efficiency remains to be seen.

\section{Discussion}
Even though our approach of function-search has significantly improved performance for ARC, we are still far from solving this problem. The performance of end-to-end deep learning methods even with extensive training does not yield decent results. Code-It \cite{codeit} achieves only 14.75\% on the evaluation set. This is related to the feature extraction capabilities of the current deep learning methods. Our CNN-based classifier also achieved only $\sim$40\% accuracy on the Concept-ARC dataset \cite{conceptarc}. Due to poor feature extraction methods, we have resorted to search-based methods guided by these approximate feature extractors. 

Effective feedback mechanisms are crucial for directing the search process in the right direction. In our approach, the scoring function served as one form of feedback alongside information on syntax errors. The scoring function provides feedback by evaluating in-context examples and assigning relative scores to guide the search. To strengthen the feedback signals, it is crucial to include information about the specific issues with in-context programs. Also, each step in the program generation process currently operates independently, which can lead to the LLM repeating previous mistakes. The only experience the system receives is derived from the evolved in-context examples. By integrating a detailed feedback mechanism and leveraging accumulated experiences, we may be able to enhance the search.

One approach tested in conjunction with function-search for addressing complex tasks in a step-by-step manner was problem-iteration, though it did not produce successful results. This method involved performing several iterations of code-iterations and then resetting the problem using the best program $f$ from the program database, as evaluated by a scoring function. In this process, the new input grid is derived as $f(I_{1:n})$, while the output grid remains unchanged. Essentially, this approach aims to solve part of the problem and then reframe the remaining portion as a new problem. However, it was observed that the best program $f$ frequently lost crucial information from the original input grid, which was necessary for reaching the desired output grid.

\section{Conclusion}
In this paper, we proposed a novel function-search algorithm incorporating a concept-based scoring method to enhance search efficiency using large language models (LLMs) for Abstraction and Reasoning Corpus (ARC). Our method achieves a notable improvement, with task performance reaching 58\% compared to the baseline's direct-grid approach, which only achieves 26\%, when evaluated on a set of 50 tasks using GPT-4. Furthermore, our concept-based scoring function demonstrates up to 30\% greater efficiency than Hamming distance, as measured by the number of code iterations needed to reach the correct solution. This advancement highlights the effectiveness of our LLM-based search strategy, which avoids the high costs associated with sampling a large number of solutions \cite{redwoodresearch}.

\bibliography{aaai25}

\begin{thebibliography}{19}
\providecommand{\natexlab}[1]{#1}

\bibitem[{Acquaviva et~al.(2023)Acquaviva, Pu, Kryven, Sechopoulos, Wong, Ecanow, Nye, Tessler, and Tenenbaum}]{larc}
Acquaviva, S.; Pu, Y.; Kryven, M.; Sechopoulos, T.; Wong, C.; Ecanow, G.~E.; Nye, M.; Tessler, M.~H.; and Tenenbaum, J.~B. 2023.
\newblock Communicating Natural Programs to Humans and Machines.
\newblock arXiv:2106.07824.

\bibitem[{Ainooson et~al.(2023)Ainooson, Sanyal, Michelson, Yang, and Kunda}]{ainooson2023neurodiversityinspiredsolverabstraction}
Ainooson, J.; Sanyal, D.; Michelson, J.~P.; Yang, Y.; and Kunda, M. 2023.
\newblock A Neurodiversity-Inspired Solver for the Abstraction \& Reasoning Corpus (ARC) Using Visual Imagery and Program Synthesis.
\newblock arXiv:2302.09425.

\bibitem[{Barke et~al.(2024)Barke, Gonzalez, Kasibatla, Berg-Kirkpatrick, and Polikarpova}]{barke2024hysynthcontextfreellmapproximation}
Barke, S.; Gonzalez, E.~A.; Kasibatla, S.~R.; Berg-Kirkpatrick, T.; and Polikarpova, N. 2024.
\newblock HYSYNTH: Context-Free LLM Approximation for Guiding Program Synthesis.
\newblock arXiv:2405.15880.

\bibitem[{Bober-Irizar and Banerjee(2024)}]{boberirizar2024neuralnetworksabstractionreasoning}
Bober-Irizar, M.; and Banerjee, S. 2024.
\newblock Neural networks for abstraction and reasoning: Towards broad generalization in machines.
\newblock arXiv:2402.03507.

\bibitem[{Butt et~al.(2024)Butt, Manczak, Wiggers, Rainone, Zhang, Defferrard, and Cohen}]{codeit}
Butt, N.; Manczak, B.; Wiggers, A.; Rainone, C.; Zhang, D.~W.; Defferrard, M.; and Cohen, T. 2024.
\newblock CodeIt: Self-Improving Language Models with Prioritized Hindsight Replay.
\newblock arXiv:2402.04858.

\bibitem[{Chollet(2019)}]{chollet2019measureintelligence}
Chollet, F. 2019.
\newblock On the Measure of Intelligence.
\newblock arXiv:1911.01547.

\bibitem[{Ellis et~al.(2020)Ellis, Wong, Nye, Sable-Meyer, Cary, Morales, Hewitt, Solar-Lezama, and Tenenbaum}]{ellis2020dreamcodergrowinggeneralizableinterpretable}
Ellis, K.; Wong, C.; Nye, M.; Sable-Meyer, M.; Cary, L.; Morales, L.; Hewitt, L.; Solar-Lezama, A.; and Tenenbaum, J.~B. 2020.
\newblock DreamCoder: Growing generalizable, interpretable knowledge with wake-sleep Bayesian program learning.
\newblock arXiv:2006.08381.

\bibitem[{Greenblatt(2024)}]{redwoodresearch}
Greenblatt, R. 2024.
\newblock Getting 50\% (SoTA) on ARC-AGI with GPT-4o.

\bibitem[{Hodel(2024)}]{arc-dsl}
Hodel, M. 2024.
\newblock ARC-DSL.

\bibitem[{Icecuber(2020)}]{icecuber}
Icecuber. 2020.
\newblock Winner of ARC Challenge.

\bibitem[{Khosla et~al.(2021)Khosla, Teterwak, Wang, Sarna, Tian, Isola, Maschinot, Liu, and Krishnan}]{khosla2021supervisedcontrastivelearning}
Khosla, P.; Teterwak, P.; Wang, C.; Sarna, A.; Tian, Y.; Isola, P.; Maschinot, A.; Liu, C.; and Krishnan, D. 2021.
\newblock Supervised Contrastive Learning.
\newblock arXiv:2004.11362.

\bibitem[{Mitchell, Palmarini, and Moskvichev(2023)}]{mitchell2023comparinghumansgpt4gpt4v}
Mitchell, M.; Palmarini, A.~B.; and Moskvichev, A. 2023.
\newblock Comparing Humans, GPT-4, and GPT-4V On Abstraction and Reasoning Tasks.
\newblock arXiv:2311.09247.

\bibitem[{Moskvichev, Odouard, and Mitchell(2023)}]{conceptarc}
Moskvichev, A.; Odouard, V.~V.; and Mitchell, M. 2023.
\newblock The ConceptARC Benchmark: Evaluating Understanding and Generalization in the ARC Domain.
\newblock arXiv:2305.07141.

\bibitem[{Reimers and Gurevych(2019)}]{reimers-2019-sentence-bert}
Reimers, N.; and Gurevych, I. 2019.
\newblock Sentence-BERT: Sentence Embeddings using Siamese BERT-Networks.
\newblock In \emph{Proceedings of the 2019 Conference on Empirical Methods in Natural Language Processing}. Association for Computational Linguistics.

\bibitem[{Romera-Paredes et~al.(2023)Romera-Paredes, Barekatain, Novikov, Balog, Kumar, Dupont, Ruiz, Ellenberg, Wang, Fawzi, Kohli, and Fawzi}]{funsearch}
Romera-Paredes, B.; Barekatain, M.; Novikov, A.; Balog, M.; Kumar, M.~P.; Dupont, E.; Ruiz, F. J.~R.; Ellenberg, J.~S.; Wang, P.; Fawzi, O.; Kohli, P.; and Fawzi, A. 2023.
\newblock Mathematical discoveries from program search with large language models.
\newblock \emph{Nature}, 625(7995): 468–475.

\bibitem[{Wang et~al.(2024)Wang, Zelikman, Poesia, Pu, Haber, and Goodman}]{wang2024hypothesissearchinductivereasoning}
Wang, R.; Zelikman, E.; Poesia, G.; Pu, Y.; Haber, N.; and Goodman, N.~D. 2024.
\newblock Hypothesis Search: Inductive Reasoning with Language Models.
\newblock arXiv:2309.05660.

\bibitem[{Xu, Khalil, and Sanner(2022)}]{arga}
Xu, Y.; Khalil, E.~B.; and Sanner, S. 2022.
\newblock Graphs, Constraints, and Search for the Abstraction and Reasoning Corpus.
\newblock arXiv:2210.09880.

\bibitem[{Xu et~al.(2024)Xu, Li, Vaezipoor, Sanner, and Khalil}]{xu2024llmsabstractionreasoningcorpus}
Xu, Y.; Li, W.; Vaezipoor, P.; Sanner, S.; and Khalil, E.~B. 2024.
\newblock LLMs and the Abstraction and Reasoning Corpus: Successes, Failures, and the Importance of Object-based Representations.
\newblock arXiv:2305.18354.

\bibitem[{Yu and Koltun(2016)}]{yu2016multiscalecontextaggregationdilated}
Yu, F.; and Koltun, V. 2016.
\newblock Multi-Scale Context Aggregation by Dilated Convolutions.
\newblock arXiv:1511.07122.

\end{thebibliography}

\end{document}